\documentclass[runningheads]{llncs}
\usepackage{graphicx}
\usepackage{amsmath}
\usepackage{amssymb}
\usepackage{hyperref}
\usepackage[absolute]{textpos}
\begin{document}
\begin{textblock}{5}(14,1)
\noindent\small This is the preliminary version. Final version has been published in The 15th Asia Information Retrieval Societies Conference proceeding (2019) and can be obtained from \url{https://link.springer.com/chapter/10.1007/978-3-030-42835-8_6}
\end{textblock}

%
\title{Context-aware Helpfulness Prediction for Online Product Reviews\thanks{The work described in this paper is substantially supported by a grant from the Research Grant Council of the Hong Kong Special Administrative Region, China (Project Code: 14204418).}}

%
%

\author{Iyiola E. Olatunji \and Xin Li \and Wai Lam}

\authorrunning{I. E. Olatunji  et al.}

%

\institute{Department of Systems Engineering and Engineering Management, \\
The Chinese University of Hong Kong, \\
Shatin, Hong Kong \\
\email{\{olatunji,lixin, wlam\}@se.cuhk.edu.hk}\\
}

\maketitle              

\begin{abstract}
Modeling and prediction of review helpfulness has become more predominant due to proliferation of e-commerce websites and online shops. Since the functionality of a product cannot be tested before buying, people often rely on different kinds of user reviews to decide whether or not to buy a product. However, quality reviews might be buried deep in the heap of a large amount of reviews. Therefore, recommending reviews to customers based on the review quality is of the essence. Since there is no direct indication of review quality, most reviews use the information that ``X out of Y'' users found the review helpful for obtaining the review quality. However, this approach undermines helpfulness prediction because not all reviews have statistically abundant votes. In this paper, we propose a neural deep learning model that predicts the helpfulness score of a review. This model is based on convolutional neural network (CNN) and a context-aware encoding mechanism which can directly capture relationships between words irrespective of their distance in a long sequence. We validated our model on human annotated dataset and the result shows that our model significantly outperforms existing models for helpfulness prediction.

\end{abstract}

\section{Introduction}

Reviews have become an integral part of user’s experience when shopping online. This trend makes product reviews an invaluable asset because they help customers make purchasing decision, consequently, driving sales~\cite{Duan2008}. Due to the enormous amount of reviews, it is important to analyze review quality and to present useful reviews to potential customers. The quality of a review can vary from a well-detailed opinion and argument, to excessive appraisal, to spam. Therefore, predicting the helpfulness of a review involves automatic detection of the influence the review will have on a customer for making purchasing decision. Such reviews should be informative and self-contained~\cite{Diaz2018}.
\begin{figure}
\begin{center}
\begin{tabular}{|p{10cm}|}
\hline \textbf{Votes: [1, 8]} \\
\textbf{HS: 0.13} \\
\textbf{HAS: 0.85} \\
{I received an updated charger as well as the updated replacement power supply due to the recall. It looks as if this has solved all of the problems. I have been using this charger for some time now with no problems. No more heat issues and battery charging is quick and accurate. I can now recommend this charger with no problem. I use this charger several times a week and much prefer it over the standard wall type chargers. I primarily use Enelope batteries with this charger.} \\
\hline
\end{tabular}
\end{center}
\caption{\label{figure-1} Example of a review text with helpfulness score. \textbf{HS} = helpfulness score based on ``X of Y approach'' while \textbf{HAS} = human annotated helpfulness score. }
\vspace{-5mm}
\end{figure}

Review helpfulness prediction have been studied using arguments~\cite{Liu2017}, aspects (ASP)~\cite{Yang2016}, structural (STR) and unigram (UGR) features~\cite{Yang2015}. Also, semantic features such as linguistic inquiry and word count (LIWC), general inquirer (INQ)~\cite{Yang2015}, and Geneva affect label coder (GALC)~\cite{Martin2014} have been used to determine review helpfulness. However, using handcrafted features is laborious and expensive due to manual feature engineering and data annotation. Recently, convolutional neural networks (CNNs)~\cite{Kim2014}, more specifically, the character-based CNN~\cite{Chen2018} has been applied to review helpfulness prediction and has shown to outperform handcrafted features. However, it does not fully capture the semantic representation of the review since different reviewers have different perspective, writing style and the reviewer’s language may affect the meaning of the review. That is, the choice of word of an experienced reviewer differs from that of a new reviewer. Therefore, modelling dependency between words is important.

Recent works on helpfulness prediction use the ``X of Y approach'' i.e. if X out of Y users votes that a review is helpful, then the helpfulness score of the review is X/Y. However, such simple method is not effective as shown in Figure~\ref{figure-1}. The helpfulness score (HS = 0.13) implies that the review is unhelpful. However, this is not the case as the review text entails user experience and provides necessary information to draw a reasonable conclusion (self-contained). Hence, it is clear that the review is of high quality (helpful) as pointed out by human annotators (HAS = 0.85). This observation demonstrates that ``X of Y'' helpfulness score may not strongly correlate to review quality~\cite{Yang2015} which undermines the effectiveness of the prediction output.

Similarly, prior methods assume that reviews are coherent ~\cite{Yang2015}\cite{Chen2018}. However, this is not the case in most reviews because of certain content divergence features such as sentiment divergence (opinion polarity of products compared to that of review) embedded in reviews. To address these issues, we propose a model that predicts the helpfulness score of a review by considering the context in which words are used in relation to the entire review. We aim to understand the internal structure of the review text.

In order to learn the internal structure of review text, we learn dependencies between words by taking into account the positional encoding of each word and employing the self-attention mechanism to cater for the length of longer sequences. We further encode our learned representation into CNN to produce an output sequence representation.

In our experiment, our framework outperforms existing state-of-the-art methods for helpfulness prediction on the Amazon review dataset. We conducted detailed experiments to quantitatively evaluate the effectiveness of each designed component of our model. We validated our model on the human annotated data. The code is available. 

\section{Related Works}
\label{sect:relatedworks}
Previous works on helpfulness prediction can be categorized broadly into three categories (a) score regression (prediction a helpfulness score between 0 and 1) (b) classification (classifying a review as either helpful or not helpful) (c) ranking (ordering reviews based on their helpfulness score). In this paper we define the problem of helpfulness prediction as a regression task. Most studies tend to focus on extracting specific features (handcrafted features) from review text.

Semantic features such as LIWC (linguistic inquiry and word count), general inquirer (INQ) ~\cite{Yang2015}, and GALC (Geneva affect label coder) ~\cite{Martin2014} were used to extract meaning from the text to determine the helpfulness of the review. Extracting argument features from review text has shown to outperform semantic features ~\cite{Liu2017} since it provides a more detailed information about the product. Structural (STR) and unigram (UGR) features~\cite{Yang2015} have also been exploited.

Content-based features such as review text and star rating and context-based features such as reviewer/user information are used for helpfulness prediction task. Content-based features are features that can be obtained from the reviews. They include length of review, review readability, number of words in review text, word-category features and content divergence features. Context-based features are features that can be derived outside the review. This includes reviewer features (profile) and features to capture similarities between users and reviews (user-reviewer Idiosyncrasy). Other metadata such as the probability of a review and its sentences being subjective have been successfully used as features ~\cite{Otterbacher:2009:HOC:1518701.1518848}\cite{Kim:2006:AAR:1610075.1610135}\cite{Mudambi:10.2307/20721420}\cite{PAN2011598}\cite{SALEHAN201630}\cite{Ipeirotis:5590249}. Since review text are mostly subjective, it is a more complicated task to model all contributing features for helpfulness prediction. Thus, using handcrafted features has limited capabilities and laborious due to manual data annotation.

Several methods have been applied to helpfulness prediction task including support vector regression ~\cite{Kim:2006:AAR:1610075.1610135,Zhang:2006:USP:1183614.1183626,Yang2015}, probabilistic matrix factorization ~\cite{Tang:2013:CRH:2507157.2507183}, linear regression ~\cite{Lu:2010:ESC:1772690.1772761}, extended tensor factorization models ~\cite{Moghaddam:2012:EET:2124295.2124316}, HMM-LDA based model ~\cite{Mukherjee:doi:10.1137/1.9781611974973.54} and multi-layer neural networks ~\cite{LEE20143041}. These methods allows the integration of robust constraints into the learning process and this in turn has improved prediction result.
Recently, convolutional neural network (CNN) has shown significant improvement over existing methods by achieving the state-of-the-art performance ~\cite{Kim2014}\cite{Zhang:2015:CCN:2969239.2969312}. CNNs automatically extract deep feature from raw text. This capability can alleviate manual selection of features. Furthermore, adding more levels of abstraction as in character-level representations has further improved prediction results over vanilla CNN. Embedding-gated CNN ~\cite{Chen2018a} and multi-domain gated CNN ~\cite{Chen:2019:MGC:3308558.3313587} are recent methods for predicting helpfulness prediction.

Moreover, attention mechanism has also been employed to CNN such as the ABCNN for modelling sentence pair~\cite{Yin2016}. Several tasks including textual entailment, sentence representation, machine translation, and abstractive summarization have applied self-attention mechanism and has shown significant result~\cite{Ambartsoumian2018}. However, employing self-attention for developing context-aware encoding mechanism has not been applied to helpfulness prediction. Using self-attention mechanism on review text is quite intuitive because even the same word can have different meaning based on the context in which it is being used.

\begin{figure}
\begin{center}
\includegraphics[width=12cm, height=7cm]{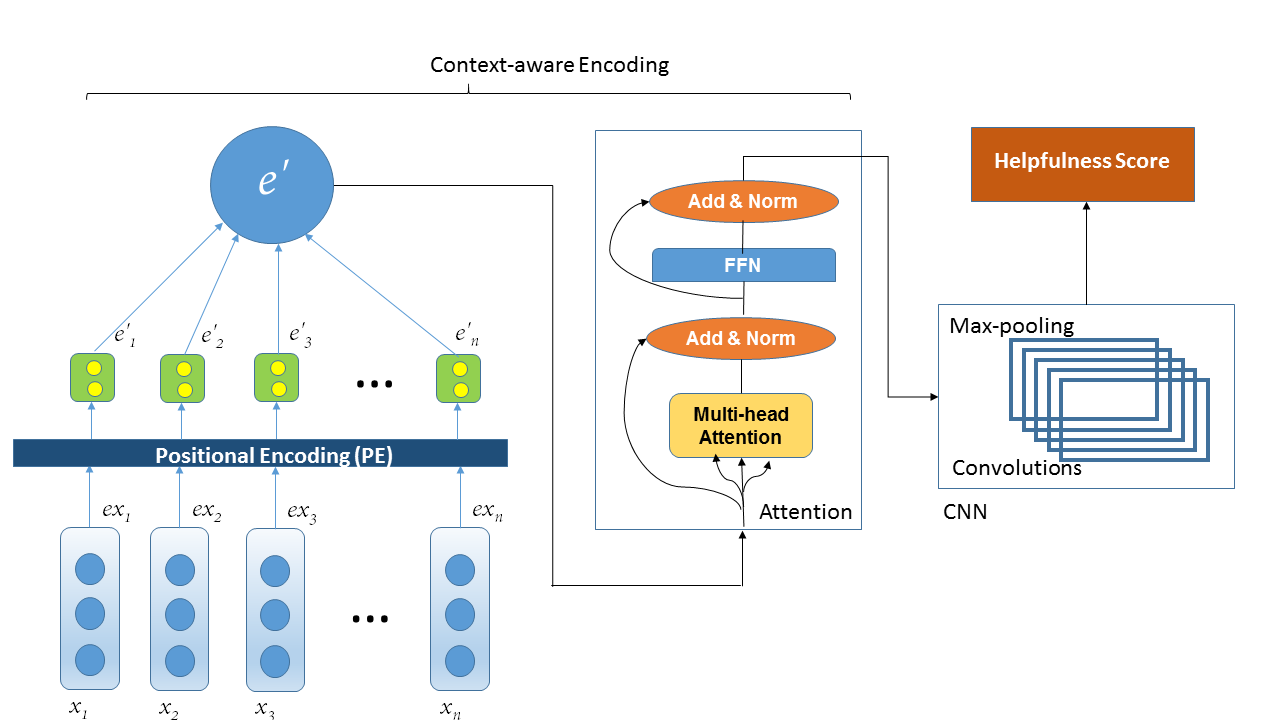}
\end{center}
\caption{\label{figure-2} Proposed context-aware helpfulness prediction model }
\end{figure}

\section{Model}
\label{sect:model}
We model the problem of predicting the helpfulness score of a review as a regression problem. Precisely, given a sequence of review, we predict the helpfulness score based on the review text. As shown in Figure~\ref{figure-2}, the sequence of words in the review are embedded and concatenated with their respective positional encoding to form the input features.
These input features are processed by a self-attention block for generating context-aware representations for all tokens. Then such representation will be fed into a convolutional neural network (CNN) which computes a vector representation of the entire sequence (i.e the dependency between the token and the entire sequence) and then we use a regression layer for predicting the helpfulness score.

\subsection{Context-aware Encoding}
\label{sect:contextencoding}
The context-aware component of our model consists of positional encoding and self-attention mechanism. We augment the word embedding with positional encoding vectors to learn text representation while taking the absolute position of words into consideration.

Let \(X = \left(x_1, x_2,...,x_n \right) \) be a review consisting of a sequence of words. We map each word \(x_i \) in a review \(X \) to a \(l \)-dimensional (word embedding) word vector \( e_{x_i} \) stored in an embedding matrix \(\textbf{E} \in {\rm I\!R}^{V\times l} \) where \( V \) is the vocabulary size. We initialize \(\textbf{E} \) with pre-trained vectors from GloVe~\cite{Pennington2014} and set the embedding dimension \(\textit{l} \) to 100.
A review is therefore represented as \(Y = \left(e_{x_1}, e_{x_2},...,e_{x_n} \right ) \). Since the above representation learns only the meaning of each word, we need the position of each word for understanding the context of each word. Let \(S = \left({s_1}, {s_2},...,{s_n} \right ) \) be the position of each word in a sentence. Inspired by Vaswani \textit{et al} \cite{Vaswani2017}, the positional encoding, denoted as \(PE \in {\rm I\!R}^{n\times l} \), is a 2D constant matrix with position specific values calculated by the sine and cosine functions below:
\begin{equation} \label{eqn1}
\begin{split}
PE \left({s_k}, 2i\right)= sin\left({s_k} / j^{2i / l} \right )  \\
PE \left({s_k}, 2i + 1 \right)= cos\left({s_k} / j^{2i / l} \right )  \\
\end{split}
\end{equation}
where \(i \) is the position along the embedding vector \(e \) and \(j \) is a constant representing the distance between successive peaks (or successive troughs) of cosine and sine function. This constant is between \(2\pi \) and 10000. Based on a tuning process, we set \(j \) to 1000.
The sequence \(\overline{P} = \left(\overline{P_1}, \overline{P_2},..., \overline{P_n} \right ) \), where \(\overline{P_s}  = PE(S) \) defined as the row vector corresponding to \(S \) in the matrix \(PE \) as in Equation \ref{eqn1}.

The final representation \(e^\prime \) is obtained by adding the word embedding to the relative position values of each word in the sequence. Therefore,
\( e^\prime = \left(e_1^\prime, e_2^\prime,...,e_n^\prime \right ) \)
where
\( e_j^\prime = \left(\overline{P_j} + e_{x_j} \right ) \)

Self-attention is employed in our model. Given an input \(e^\prime \), self-attention involves applying the attention mechanism on each \(\mathit{e_i^\prime} \) using \(\mathit{e_i^\prime} \) query vector and key-value vector for all other positions.
The reason for using the self-attention mechanism is to capture the internal structure of the sentence by learning dependencies between words. The scaled dot-product attention is used which allows faster computation instead of the standard additive attention mechanism~\cite{Bahdanau2015}. It computes the attention scores by:
\begin{equation} \label{eqn4}
Attention(Q,K,V) = softmax (\frac{(Q)(K)^T}{\sqrt{l}})V
\end{equation}
where \(\mathit{Q, K} \) and \(\mathit{V} \)  are the query, key and value matrices respectively.
The above equation implies that we divide the dot product of queries with all keys by the key vector dimension to obtain the weight on the values.

We used multi-head attention similar to~\cite{Vaswani2017}. The multi-head attention mechanism maps the input vector \(\mathit{e^\prime} \) to queries, keys and values matrices by using different linear projections. This strategy allows the self-attention mechanism to be applied \(\mathit{h} \) times. Then all the vectors produced by different heads are concatenated together to form a single vector.

Concisely, our model captures context for a sequence as follows: We obtain the relative position of the tokens in the sequence from Equation \ref{eqn1}.
The self attention block, learns the context by relating or mapping different positions \(s_1, s_2, ..., s_n \) of \(\overline{P} \) via Equation \ref{eqn1} so as to compute a single encoding representation of the sequence. By employing the multi-head attention, our model can attend to words from different encoding representation at different positions. We set heads \(\textit{h} \) to 2.
The query, key and value used in the self-attention block are obtained from the output of the previous layer of the context-aware encoding block. This design allows every position in the context-aware encoding block to attend over all positions of the input sequence.
To consider positional information in longer sentences not observed in training, we apply the sinusoidal position encoding  to the input embedding.

\subsection{Convolutional Neural Network (CNN)}
The output of the context-aware encoding representation \(e^\prime \) is fed into the CNN to obtain new feature representations for making predictions. We employ multiple filters \(f \in \) [1,2,3]. This method is similar to learning uni-gram, bi-gram and tri-gram representations respectively. Specifically, for each filter, we obtain an hidden representation \(r = Pool(Conv(e^\prime, filterSize(f,l,c))) \) where \(c\) is the channel size, \(Pool\) is the pooling operation and \(Conv(.)\) is the convolution operation. In our experiment, we use max pooling and average pooling. The final representation \( h \) is obtained by concatenating all hidden representation. i.e., \( h  = [r_1, r_2, r_3]\). These features are then passed to the regression layer to produce the helpfulness scores.

\section{Dataset}
\label{sect:dataset}
We used two datasets for our experiments. The first dataset called D1 is constructed from the Amazon product review dataset~\cite{Mcauley2015}. This dataset consists of over 142 million reviews from Amazon between 1996 to 2014. We used a subset of 21,776,678 reviews from 5 categories, namely; health, electronics, home, outdoor and phone. We selected reviews with over 5 votes as done in~\cite{Chen2018,Yang2015}. The statistics of the dataset used are shown in table \ref{dataset-table}. We removed reviews having less than 7 words for experimenting with different filter sizes. Note that this is the largest dataset used for helpfulness prediction task.

The second dataset called D2 is the human annotated dataset from Yang \textit{et al.}~\cite{Yang2015}. This dataset consists of 400 reviews with 100 reviews selected randomly from four product categories (outdoor, electronics, home and books). The reason for using the human annotated dataset is to verify that our model truly learns deep semantics features of review text. Therefore, our model was not trained on the human annotated dataset but only used for evaluating the effectiveness of our model. We used only three categories for our experiment and performed cross-domain experiment on categories not in D2.

\begin{table}
\caption{\label{dataset-table} Data statistic of Amazon reviews from 5 different categories. We used Health instead of Watches as done by Chen \textit{et al.} ~\cite{Chen2018} because it is excluded from newly published Amazon dataset.  }
\begin{center}
\begin{tabular}{|p{3cm}|p{5cm}|p{3cm}|}
\hline \textbf{Product category} & \textbf{\# of reviews with 5+ votes} & \textbf{Total \# of reviews} \\ 
\hline Phone & 261,370 & 3,447,249 \\ 
\hline Outdoor & 491,008 & 3,268,695 \\
\hline Health & 550,297 & 2,982,326 \\
\hline Home & 749,564 & 4,253,926 \\
\hline Electronics & 1,310,513 & 7,824,482 \\
\hline
\end{tabular}
\end{center}
\end{table}

\section{Experimental Setup}
\label{sect:experiment}
Following the previous works~\cite{Chen2018,Yang2015,Yang2016}, all experiments were evaluated using correlation coefficients between the predicted helpfulness scores and the ground truth scores. We split the dataset D1 into Train / Test / Validation (70, 20, 10). We used the same baselines as state-of-the-art convolutional model for helpfulness prediction~\cite{Chen2018} i.e. STR, UGR, LIWC, INQ~\cite{Yang2015}, ASP~\cite{Yang2016}. CNN is the CNN model by~\cite{Kim2014} and C\_CNN is the state-of-the-art character-based CNN from~\cite{Chen2018}. We added two additional variants (S\_Attn and S\_Avg) to test different components of our model.
S\_Attn involves using only self-attention without CNN while S\_Avg is self-attention with CNN using average pooling and finally our model uses max pooling with context-aware encoding.
We re-implemented all baselines as well as C\_CNN as described by \cite{Chen2018} but excluded the transfer learning part of their model since it is for tackling insufficient data problem. We used RELU for non-linearity and set dropout rate to 0.5 (for regularization). We used Adaptive moment estimation (Adam) as our optimizer. The learning rate was set to 0.001 and \(l \) to 100. We experimented with different filter sizes and found that \(f \in \) [1,2,3] produces the best result. Also we tried using Recurrent Neural Network (RNN) such as LSTM and BiLSTM but they performed worse than CNN.

\begin{table}
\caption{\label{exisres-table} Experimental result for the dataset D1 }
\begin{center}
\begin{tabular}{|p{1.2cm}|p{1.5cm}|p{1.7cm}|p{1.5cm}|p{1.5cm}|p{2cm}|}
\hline \textbf{} & \textbf{Phone} & \textbf{Outdoor} &\textbf{Health} & \textbf{Home} & \textbf{Electronics} \\ 
\hline STR & 0.136 & 0.210 & 0.295 & 0.210 & 0.288 \\
\hline UGR & 0.210 & 0.299 & 0.301 & 0.278 & 0.310 \\
\hline LIWC & 0.163 & 0.287 & 0.268 & 0.285 & 0.350 \\
\hline INQ & 0.182 & 0.324 & 0.310 & 0.291 & 0.358 \\
\hline ASP & 0.185 & 0.281 & 0.342 & 0.233 & 0.366 \\
\hline CNN & 0.221 & 0.392 & 0.331 & 0.347 & 0.411 \\
\hline C\_CNN & 0.270 & 0.407 & 0.371 & 0.366 & 0.442 \\
\hline S\_Attn & 0.219 & 0.371 & 0.349 & 0.358 & 0.436 \\
\hline S\_Avg & 0.194 & 0.236 & 0.336 & 0.318 & 0.382 \\
\hline Ours & \textbf{0.373} & \textbf{0.461} & \textbf{0.428} & \textbf{0.402} & \textbf{0.475} \\
\hline
\end{tabular}
\end{center}
\end{table}

\begin{table}
\caption{\label{humananot-table} Experimental result for the dataset D2. (Fusion\_all = STR + UGR + LIWC + INQ) }
\begin{center}
\begin{tabular}{|p{1.7cm}|p{1.5cm}|p{1.5cm}|p{2cm}|}
\hline \textbf{} & \textbf{Outdoor} & \textbf{Home} &\textbf{Electronics} \\ 
\hline Fusion\_all & 0.417 & 0.596 & 0.461 \\
\hline CNN & 0.433 & 0.521 & 0.410 \\
\hline C\_CNN & 0.605 & 0.592 & 0.479 \\
\hline Ours & \textbf{0.748} & \textbf{0.758} & \textbf{0.699} \\
\hline
\end{tabular}
\end{center}
\end{table}

\begin{table}
\caption{\label{crossdom-table} Cross-domain investigation }
\begin{center}
\begin{tabular}{|p{1.9cm}|p{3cm}|p{3cm}|}
\hline  & D1-Phone \newline D2-Home  & D1-Health \newline D2-Electronics \\
\hline C\_CNN         & 0.389 & 0.436  \\
\hline Ours & \textbf{0.586} & \textbf{0.654} \\
\hline
\end{tabular}
\end{center}

\end{table}

\section{Results}
\label{sect:results}
As shown in Table~\ref{exisres-table}, our context-aware encoding based model using max pooling outperforms all handcrafted features and CNN-based models including C\_CNN with a large margin on D1. This is because by applying attention at different positions of the word embedding, different information about word dependencies are extracted which in turn handles context variation around the same word. However, using self-attention alone (S\_Attn) (table~\ref{exisres-table}) performs poorly than CNN as learning word dependencies alone is not sufficient for our task. We further need to understand the internal structure of the review text.
Since self-attention can handle longer sequence length than CNN when modelling dependencies, we resolve to capturing the dependencies using self-attention and then encode the dependencies into a vector representation using CNN to further extract the positional invariant features.
Two variants are presented using average pooling (S\_Avg) and max pooling.
S\_Avg performs comparable to handcrafted features probably due to its tendency of selecting tokens having low attention scores.
Our proposed model with max-pooling produces the best result on D1 (Table~\ref{exisres-table}) and significantly on D2 (Table~\ref{humananot-table}) since it selects the best representation with most attention. It implies that our model can capture the dependency between tokens and the entire sequence. Likewise, our model understands the internal structure of review and has a high correlation to human score.

Since D2 does not include the Phone and Health category, we tested our proposed model trained on the Phone and Health category from D1 on the Home and Electronics category respectively on D2.
Specifically, we used the training data of the Phone category from D1 to train our proposed model and used the data of the Home category from D2 for testing. Similarly, we used the training data of the Health category from D1 to train our proposed model and tested the model using the data from the Electronics category of D2.

As shown in Table~\ref{crossdom-table}, the result is quite surprising. This shows that our proposed model can effectively learn cross domain features and is robust to ``out-of-vocabulary'' (OOV) problem by predicting reasonable helpfulness score having a high correlation to human score.

\section{Conclusions}
\label{sect:conclusion}

Predicting review helpfulness can substantially save a potential customer’s time by presenting the most useful review. 
In this paper, we propose a context-aware encoding based method that learns dependencies between words for understanding the internal structure of the review.Experimental results on the human annotated data shows that our model is a good estimator for predicting the helpfulness of reviews and robust to the “out-of-vocabulary” (OOV) problem. In the future, we aim to explore some learning to rank models to effectively rank helpfulness score while incorporating some other factors that may affect helpfulness prediction including the types of products.


\bibliographystyle{splncs04}
\bibliography{ref}

\begin{thebibliography}{10}
\providecommand{\url}[1]{\texttt{#1}}
\providecommand{\urlprefix}{URL }
\providecommand{\doi}[1]{https://doi.org/#1}

\bibitem{Ambartsoumian2018}
Ambartsoumian, A., Popowich, F.: {Self-Attention : A Better Building Block for
  Sentiment Analysis Neural Network Classifiers}. In: Proceedings of the 9th
  Workshop on Computational Approaches to Subjectivity, Sentiment and Social
  Media Analysis. pp. 130--139 (2018)

\bibitem{Bahdanau2015}
Bahdanau, D., Cho, K., Bengio, Y.: {Neural machine translation by jointly
  learning to align and translate}. In: Proc. of ICLR. pp. 1--15 (2015)

\bibitem{Chen2018a}
Chen, C., Qiu, M., Yang, Y., Zhou, J., Huang, J., Li, X., Bao, F.S.: {Review
  Helpfulness Prediction with Embedding-Gated CNN}. arXiv  (2018)

\bibitem{Chen:2019:MGC:3308558.3313587}
Chen, C., Qiu, M., Yang, Y., Zhou, J., Huang, J., Li, X., Bao, F.S.:
  Multi-domain gated cnn for review helpfulness prediction. In: Proc. of WWW.
  pp. 2630--2636 (2019)

\bibitem{Chen2018}
Chen, C., Yang, Y., Zhou, J., Li, X., Bao, F.S.: {Cross-Domain Review
  Helpfulness Prediction based on Convolutional Neural Networks with Auxiliary
  Domain Discriminators}. In: Proc. of NAACL-HLT. pp. 602--607 (2018)

\bibitem{Diaz2018}
Diaz, G.O., Ng, V.: {Modeling and Prediction of Online Product Review
  Helpfulness : A Survey}. In: Proc. of ACL. pp. 698--708 (2018)

\bibitem{Duan2008}
Duan, W., Gu, B., Whinston, A.B.: {The dynamics of online word-of-mouth and
  product sales — An empirical investigation of the movie industry}. Journal
  of Retailing  \textbf{84},  233--242 (2008)

\bibitem{Ipeirotis:5590249}
{Ghose}, A., {Ipeirotis}, P.G.: Estimating the helpfulness and economic impact
  of product reviews: Mining text and reviewer characteristics. IEEE
  Transactions on Knowledge and Data Engineering  \textbf{23}(10),  1498--1512
  (2011)

\bibitem{Kim:2006:AAR:1610075.1610135}
Kim, S.M., Pantel, P., Chklovski, T., Pennacchiotti, M.: Automatically
  assessing review helpfulness. In: Proc. of EMNLP. pp. 423--430 (2006)

\bibitem{Kim2014}
Kim, Y.: {Convolutional Neural Networks for Sentence Classification}. In: Proc.
  of EMNLP. pp. 1746--1751 (2014)

\bibitem{LEE20143041}
Lee, S., Choeh, J.Y.: Predicting the helpfulness of online reviews using
  multilayer perceptron neural networks. Expert Systems with Applications
  \textbf{41}(6),  3041 -- 3046 (2014)

\bibitem{Liu2017}
Liu, H., Gao, Y., Lv, P., Li, M., Geng, S., Li, M., Wang, H.: {Using
  Argument-based Features to Predict and Analyse Review Helpfulness}. In: Proc.
  of EMNLP. pp. 1358--1363 (2017)

\bibitem{Lu:2010:ESC:1772690.1772761}
Lu, Y., Tsaparas, P., Ntoulas, A., Polanyi, L.: Exploiting social context for
  review quality prediction. In: In Proc. of WWW. pp. 691--700 (2010)

\bibitem{Martin2014}
Martin, L., Pu, P.: {Prediction of Helpful Reviews Using Emotions Extraction}.
  In: Proc. of AAAI. pp. 1551--1557 (2014)

\bibitem{Mcauley2015}
Mcauley, J., Targett, C., Hengel, A.V.D.: {Image-based Recommendations on
  Styles and Substitutes}. In: Proc. of SIGIR (2015)

\bibitem{Moghaddam:2012:EET:2124295.2124316}
Moghaddam, S., Jamali, M., Ester, M.: Etf: Extended tensor factorization model
  for personalizing prediction of review helpfulness. In: Proceedings of the
  Fifth ACM International Conference on Web Search and Data Mining. pp.
  163--172 (2012)

\bibitem{Mudambi:10.2307/20721420}
Mudambi, S.M., Schuff, D.: Research note: What makes a helpful online review? a
  study of customer reviews on amazon.com. MIS Quarterly  \textbf{34}(1),
  185--200 (2010)

\bibitem{Mukherjee:doi:10.1137/1.9781611974973.54}
Mukherjee, S., Popat, K., Weikum, G.: Exploring latent semantic factors to find
  useful product reviews. In: Proceedings of the 2017 SIAM International
  Conference on Data Mining. pp. 480--488 (2017)

\bibitem{Otterbacher:2009:HOC:1518701.1518848}
Otterbacher, J.: 'helpfulness' in online communities: A measure of message
  quality. In: Proceedings of the SIGCHI Conference on Human Factors in
  Computing Systems. pp. 955--964 (2009)

\bibitem{PAN2011598}
Pan, Y., Zhang, J.Q.: Born unequal: A study of the helpfulness of
  user-generated product reviews. Journal of Retailing  \textbf{87}(4),  598 --
  612 (2011)

\bibitem{Pennington2014}
Pennington, J., Socher, R., Manning, C.D.: {GloVe : Global Vectors for Word
  Representation}. In: Proc. of EMNLP. pp. 1532--1543 (2014)

\bibitem{SALEHAN201630}
Salehan, M., Kim, D.J.: Predicting the performance of online consumer reviews:
  A sentiment mining approach to big data analytics. Decision Support Systems
  \textbf{81},  30 -- 40 (2016)

\bibitem{Tang:2013:CRH:2507157.2507183}
Tang, J., Gao, H., Hu, X., Liu, H.: Context-aware review helpfulness rating
  prediction. In: Proceedings of the 7th ACM Conference on Recommender Systems
  (2013)

\bibitem{Vaswani2017}
Vaswani, A., Shazeer, N., Parmar, N., Uszkoreit, J., Jones, L., Gomez, A.N.,
  Kaiser, {\L}., Polosukhin, I.: {Attention Is All You Need}. In: Proc. of NIPS
  (2017)

\bibitem{Yang2016}
Yang, Y., Chen, C., Bao, F.S.: {Aspect-Based Helpfulness Prediction for Online
  Product Reviews}. In: Proc. of International Conference on Tools with
  Artificial Intelligence (ICTAI). pp. 836--843 (2016).
  \doi{10.1109/ICTAI.2016.0130}

\bibitem{Yang2015}
Yang, Y., Yan, Y., Qiu, M., Bao, F.S.: {Semantic Analysis and Helpfulness
  Prediction of Text for Online Product Reviews}. In: Proc. of ACL-IJCNLP. pp.
  38--44 (2015)

\bibitem{Yin2016}
Yin, W., Schutze, H., Xiang, B., Zhou, B.: {ABCNN : Attention-Based
  Convolutional Neural Network for Modeling Sentence Pairs}. Transactions of
  the Association for Computational Linguistics  \textbf{4},  259--272 (2016)

\bibitem{Zhang:2015:CCN:2969239.2969312}
Zhang, X., Zhao, J., LeCun, Y.: Character-level convolutional networks for text
  classification. In: Proc. of NIPS. pp. 649--657 (2015)

\bibitem{Zhang:2006:USP:1183614.1183626}
Zhang, Z., Varadarajan, B.: Utility scoring of product reviews. In: Proceedings
  of the 15th ACM International Conference on Information and Knowledge
  Management. pp. 51--57 (2006)

\end{thebibliography}

\end{document}